%% file: acl2023.tex
\pdfoutput=1
\documentclass[11pt]{article}

\usepackage{ACL2023}

\usepackage{times}
\usepackage{latexsym}

\usepackage[T1]{fontenc}
\usepackage[utf8]{inputenc}

\usepackage{microtype}

\usepackage{inconsolata}

\usepackage{booktabs}
\usepackage{listings}
\usepackage{courier}
\usepackage{graphicx}
\usepackage{amssymb}
\usepackage{hyperref}
\usepackage{algorithm} % For the floating algorithm environment
\usepackage{algpseudocode} % For pseudocode commands like \State, \If, etc.
\usepackage{pgf-pie}
\usepackage{xcolor}
\usepackage{geometry}
\usepackage{enumitem}

\title{Controllable Conversational Theme Detection Track at DSTC 12}

\author{
  Igor Shalyminov, Hang Su, Jake Vincent, Siffi Singh, Jason Cai, James Gung, Raphael Shu, Saab Mansour \\
  Amazon \\
  {\small \texttt{\{shalymin,shawnsu,jakevinc,siffis,cjinglun,gungj,zhongzhu,saabm\}@amazon.com}}
}

\begin{document}
\maketitle
\begin{abstract}
Conversational analytics has been on the forefront of transformation driven by the advances in Speech and Natural Language Processing techniques. Rapid adoption of Large Language Models (LLMs) in the analytics field has taken the problems that can be automated to a new level of complexity and scale.

In this paper, we introduce \textit{Theme Detection} as a critical task in conversational analytics, aimed at automatically identifying and categorizing topics within conversations. This process can significantly reduce the manual effort involved in analyzing expansive dialogs, particularly in domains like customer support or sales. Unlike traditional dialog intent detection, which often relies on a fixed set of intents for downstream system logic, themes are intended as a direct, user-facing summary of the conversation's core inquiry. This distinction allows for greater flexibility in theme surface forms and user-specific customizations.

We pose \textit{Controllable Conversational Theme Detection} problem as a public competition track at Dialog System Technology Challenge (DSTC) 12~--- it is framed as joint clustering and theme labeling of dialog utterances, with the distinctive aspect being controllability of the resulting theme clusters' granularity achieved via the provided user preference data.

We give an overview of the problem, the associated dataset and the evaluation metrics, both automatic and human. Finally, we discuss the participant teams' submissions and provide insights from those. The track materials (data and code) are openly available in the \href{https://github.com/amazon-science/dstc12-controllable-conversational-theme-detection}{GitHub repository}.
\end{abstract}

\section{Introduction}
\label{sec:intro}

\input{intro.tex}

\section{Related Work}
\label{sec:related}

\input{related}

\section{Task Description}
\label{sec:task}
\input{task.tex}

\section{Data}
\label{sec:data}
\input{data.tex}

\section{Baseline and Experimental Setup}
\label{sec:setup}
\input{setup.tex}

\section{Evaluation}
\label{sec:eval}
\input{eval.tex}

\section{Results and Analysis}
\label{sec:results}
\input{results}

\section{Conclusions}
\label{sec:conclusions}
\input{conclusions}

\section{Acknowledgements}
\label{sec:acknowledgements}
We express our deep gratitude to Dr. Daniel Goodhue for his assistance at the final submission evaluation stage.

\bibliography{anthology,custom}
\bibliographystyle{acl_natbib}

\appendix

\section{Cluster Labeling Prompt}
\label{sec:apx-theme-labeling-prompt}
\input{apx_theme_labeling_prompt}

\section{Theme Label Writing Guideline}
\label{sec:apx-theme-labeling-guideline}
\input{apx_theme_labeling_guideline}

\section{Human Evaluation Guidelines}
\label{sec:apx-human-eval-guidelines} 
\input{apx_human_eval_guidelines}

\section{Input/Output Data Examples}
\label{sec:apx-io-examples}
\input{apx_io_examples}

\end{document}

%% file: intro.tex
Conversational analytics~--- at the intersection of Speech and Natural Language Processing~--- has undergone rapid transformation due to advances in both fields. \textit{Automatic Speech Recognition (ASR)} now enables accurate transcription of conversations across diverse domains and durations. Simultaneously, \textit{Natural Language Processing} (especially \textit{Information Retrieval}) has enabled large-scale analysis of conversational data, revealing patterns such as word usage, emotional tone, and discussed topics. More recently, \textit{Large Language Models (LLMs)} have elevated the complexity and quality of analysis tasks. For instance, large-scale text embedding models \cite{wang-etal-2024-improving-text} significantly enhance document similarity search by capturing semantic meaning beyond surface forms.

In this paper, we propose the task of \textit{Theme Detection}, a key problem in conversational analytics. Themes reflect the high-level topics discussed in conversations and aid in categorizing them by function~--- e.g., customer support, sales, or marketing. Automatically identifying and labeling themes can greatly reduce the manual effort required to analyze long conversations.

While related to dialog intent detection, theme detection serves a different purpose. Intents are typically tied to a fixed schema and used for downstream system logic. In contrast, themes are final outputs for users (e.g., analysts), summarizing the customer's inquiry and supporting diverse surface forms and customizations.

We introduce the task of \textit{Controllable Conversational Theme Detection} as a new track in the Dialog System Technology Challenge (DSTC) 12. Building on the DSTC 11 track on Open Intent Induction \cite{gung-etal-2023-intent}, our challenge adds two major innovations: (1) joint theme detection and labeling, and (2) controllable theme granularity. The latter enables customization of theme clusters based on user preferences~--- motivated by real-world use cases where businesses may want finer or coarser thematic distinctions.

This task is designed for a zero-shot setting on unseen domains. Models will be guided by user preference data (detailed in Section~\ref{sec:data}) to align both labels and cluster granularity. While especially compelling in the context of LLMs, the proposed setup does not require their use.

%% file: related.tex
In this section, we discuss prior work related to the distinctive aspects of our proposed task.

\subsection{Unsupervised dialog theme / intent detection}
The task of open conversational intent induction was introduced in a DSTC 11 track by \citet{gung-etal-2023-intent}, which focused on utterance clustering in two setups of varying complexity: (1) intent detection with pre-defined intentful utterances to be clustered, and (2) open intent induction, which required identifying and clustering such utterances.

In contrast, our task involves a single setup with pre-defined themed utterances, and the goal is to jointly cluster and label them according to specific evaluation metrics. Unlike intent induction, we do not restrict the surface form of theme labels. Instead, labels are assessed based on their structural quality and functional usefulness for analysis (see Section~\ref{sec:eval} and Appendix~\ref{sec:apx-theme-labeling-guideline}).

\subsection{Controllable clustering}
Our goal of controllable theme granularity builds on the concept of \textit{constrained clustering}. A comprehensive taxonomy of constraint-based clustering tasks is provided by \citet{DBLP:journals/air/GonzalezAlmagroPPCG25}. We adopt instance-level pairwise constraints (``should-link'' / ``cannot-link''), implementing a semi-supervised clustering approach where supervision comes from labeled utterance pairs. This setup has been well-studied, from early work by \citet{DBLP:conf/sdm/BasuBM04} to more recent approaches by \citet{DBLP:conf/pkdd/ZhangBD19} and \citet{viswanathan-etal-2024-large}.

\subsection{Clustering with LLMs}
The use of LLMs for utterance clustering has gained traction. \citet{zhang-etal-2023-clusterllm} propose using hard triplets (``does A match B better than C?'') derived from a teacher LLM to fine-tune a smaller embedding model and refine clusters via a hierarchical method similar to HAC~\cite{Manning_Raghavan_Schütze_2008}. While this method enables controllable clustering guided by LLMs, it focuses solely on clustering~--- cluster labeling remains out of scope. In contrast, our task requires \textit{labeled theme clusters}, combining clustering with label generation to better reflect real-world needs.

\citet{viswanathan-etal-2024-large} provide a thorough study on integrating LLMs into clustering workflows. They identify three points of intervention: (1) \textit{pre-clustering}, using LLMs to generate keywords and enrich input texts; (2) \textit{during clustering}, by expanding human-provided pairwise constraints; and (3) \textit{post-clustering}, correcting uncertain assignments with LLM-based prompting. Although their framework aligns well with our goals, their focus remains on clustering rather than labeling.

%% file: task.tex
\begin{figure*}[t]
  \centering
  \includegraphics[width=\textwidth]{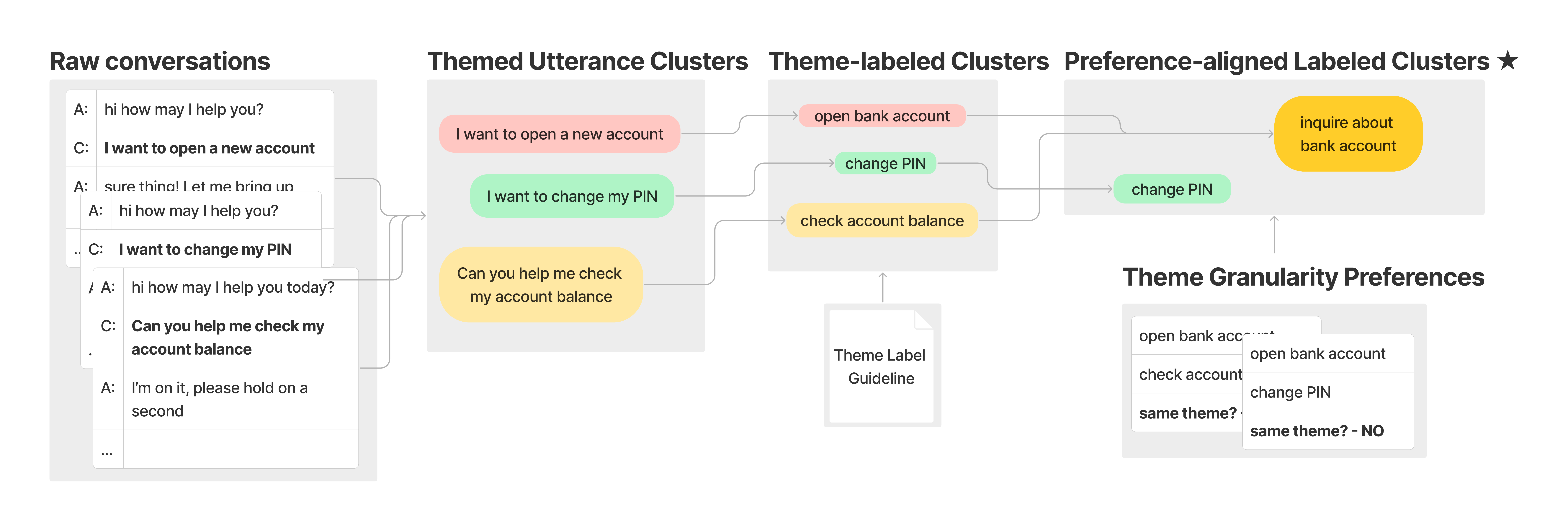}
  \caption{Diagram of the proposed task in the form of an example processing pipeline. The inputs to the ``system'' are raw conversations, user preferences on the theme granularity and theme label guidelines; the output is preference-aligned utterance clusters with the corresponding theme labels (marked with $\bigstar$)}
  \label{fig:task}
\end{figure*}

The task of \textit{Controllable Conversational Theme Detection} is defined as follows. 
The input data are:
\begin{enumerate}
 \item a dataset of conversations with some utterances within them labeled as ``themed'' (those conveying the customer's requests, possibly several per conversation)
 \item a set of preference pairs covering a sample of all the themed utterances and representing what pairs should belong to the same theme and which should not.~--- which we refer to as ``should-link'' and ``cannot-link'' pairs, respectively
 \item a theme label writing guideline outlining the requirements to a label as both a linguistic expression and an analytical tool.
\end{enumerate}

The goal of the task is to:
\begin{itemize}
 \item cluster the themed utterances so that each cluster represents a meaningful semantic / thematic group, is distinguishable from other theme clusters and satisfies the should-link~/ cannot-link requirements on its utterances (if it contains utterances included in the preference data)
 \item give each theme a short, concise and actionable natural language label (more detail on our evaluation criteria is given in Section \ref{sec:eval}).
\end{itemize}

\subsection{Controlling theme granularity}
\label{sec:theme_granularity}
In the way we intend to control theme granularity, we loosely follow the Stage 2 approach of \citealt{zhang-etal-2023-clusterllm}. That work described a data-efficient 
approach with user preference data in the should-link~/ cannot-link form. As such, if user preferences indicate that the utterances \textit{``I want to purchase pet insurance''} and \textit{``I want to purchase travel insurance''} should belong to the same theme, all the utterances like these two would be associated to the single theme whose label semantically unifies both of the two utterances' meanings e.g. \textit{``purchase insurance''} or some close paraphrase of it. On the other hand, if the preferences elicit that \textit{``I want to find the closest branch''} and \textit{``Give me the directions to the closest ATM''} should not belong to the same theme, the corresponding themes \textit{``find branch''} and \textit{``find ATM''} as well as the clusters of utterances belonging to them should be kept as separate. Some example usages of such data include contrastive fine-tuning of utterance representation as done by e.g. \citet{chu-etal-2023-multi} and \citet{zhang-etal-2021-supporting} or  adjusting the initial clusters/themes, as depicted in Figure \ref{fig:task}.

\subsection{Expected result}
\label{ref:solution}
A successful completion of the task would assume assigning each utterance a theme label so that:
\begin{itemize}
    \item theme labels are concise, exhaustively cover all the examples and are mutually exclusive,
    \item label wording conforms to the Theme label writing guideline (Appendix \ref{sec:apx-theme-labeling-guideline}),
    \item theme granularity matches the `gold' held-out assignment which is supposed to be inferred from the provided user preference samples.
\end{itemize}

A visualization of the overall task is presented in Figure \ref{fig:task} where we depict a potential sequential pipeline as an example. The actual submissions can vary in architecture and the types of models used. We intend the problem to be solved in a zero-shot weakly supervised way, in the sense that all the training/development data provided to the participants has no domain overlap with the test data (more detail on the data in Section \ref{sec:data}), and the only supervision signals provided are 1) user preference data covering a sample of the dataset and 2) theme label writing guideline. 

While the input data suggests LLM-based solutions, we encourage the participants to use techniques from both LLM-based and traditional Machine Learning paradigms that adequately correspond to the problem specifics.

\begin{table*}[ht!]
    \centering
    \small
    \caption{User Preference Data Statistics}
    \label{tab:preference-data-stats}
    \begin{tabular}{lrrrrr}
        \toprule
        \textbf{Domain} & \textbf{\# Should-link pairs} & \textbf{\% data covered} & \textbf{\# Cannot-link pairs} & \textbf{\% data covered} \\
        \midrule
        Banking & 164 & 10.04\% &  164 & 10.04\% \\
        Finance & 173 & 10\% & 173 & 10\% \\
        Insurance & 155 & 8.99\% & 126 & 7.30\% \\
        Travel (held out) & 77 & 10.07\% & 76 & 9.93\% \\
        \bottomrule
    \end{tabular}
\end{table*}

%% file: data.tex
\begin{table}
    \small
    \caption{Dialog Dataset Statistics}
    \label{tab:data-stats}
    \begin{tabular}{lrr}
        \toprule
        \textbf{Domain} & \textbf{\# Dialogs} & \textbf{\# Themed utterances} \\
        \midrule
        Banking & 980 & 1634 \\
        Finance & 3000 & 1725 \\
        Insurance & 836 & 1333 \\
        Travel (held out) & 999 & 765 \\
        \bottomrule
    \end{tabular}
\end{table}

We build our task on top of the {N}at{CS} \cite{gung-etal-2023-natcs,gung-etal-2023-intent}, a multi-domain dataset of human-human customer support conversations~--- the dataset statistics per domain are provided in Table \ref{tab:data-stats}.

We intend for the participants' submissions to work in a zero-shot setup naturally supported within the LLM-centered framework. As such, we provide the three original NatCS domains: {\bf Banking}, {\bf Finance} and {\bf Insurance}~--- for the participants to use for the training/development purposes and assess the domain generalization of their approaches.

Our theme labels closely resemble the original intent annotations in NatCS, though those were altered in the following ways:
\begin{enumerate}
 \item intent labels' surface form was rewritten where needed to conform with the theme label writing guideline (see Appendix \ref{sec:apx-theme-labeling-guideline}),
 \item for each original intent label, we provide two theme labels, a more specific one and a more vague one, for the flexibility of evaluation,
 \item intent clustering itself was altered to reflect our task's custom theme granularity,
 \item some noisy intent annotations were corrected or otherwise dropped.
\end{enumerate}

The held out test domain, {\bf Travel}, is publicly released for the first time in this challenge and has little to no overlap with the train/dev data.

Also introduced in this challenge is theme granularity preference data on top of NatCS dialogs, its statistics are shown in Table \ref{tab:preference-data-stats}. We generated preference pairs in the following way.

\textbf{Should-link pairs}: we clustered themed utterances (we leave the specifics of the clustering algorithm behind to prevent evaluation metric hacking) and sampled pairs that belong to the same cluster in the gold assignment but to the different clusters as per the algorithm, with sampling weights set to the normalized cosine distances between the points in the pair (further points that should be in the same theme are more interesting).
\textbf{Cannot-link pairs}: similarly, we sampled pairs of utterances that belong to different clusters in the gold assignment but to the same cluster as per the algorithm. Sample weights set to $1 - dist(utt_a, utt_b)$ normalized to make a probability distribution, where $utt_a$ and $utt_b$ are the utterances in the pair and $dist$ is cosine distance.

In each case, our target amount of pairs to generate corresponds to 10\% of all the themed utterances in the dataset, and preference pairs cover no more than 30\% of any given gold cluster's utterances.

%For the train and dev domains, we will provide the following data:
%
%\begin{enumerate}
%    \item Utterances to cluster/label with dialogue contexts~--- since some of the valid themed utterances might be elliptical/have otherwise incomplete explicit information, we will provide full dialogue contexts along with them. The utterances marked as themed would be the datapoints to run prediction on. See also Appendix \ref{sec:apx_data_examples} for an example input.
%    \item User preferences for clustering~--- a set of utterance pairs (with the corresponding dialogue contexts) along with binary decisions whether they should belong to the same cluster/theme or not. This works as the main input for inferring the desired theme granularity. See also Appendix \ref{sec:apx_data_examples} for a user preference example.
%    \item The `gold' theme labeling for all the themed utterances.
%\end{enumerate}
%
%For the final evaluation phase, we will release the themed utterances with dialogue contexts and the %user preferences set for the Travel domain; the gold theme labeling will be held hidden.
%
%In addition, all the train/dev/test domains will share the same theme label writing guideline (Appendix \ref{sec:apx_theme_label_guideline}) that will be available for the participants from the challenge's start.
%
%We are conducting research on the optimal number of user preference samples to sufficiently describe the desired granularity while maintaining the challengingness of the task.

%% file: setup.tex
We provided the participants with a baseline solution that combines traditional machine learning approaches with LLM-based techniques. As such, the entire baseline workflow consists of 3 stages:
\begin{enumerate}
 \item \textbf{Utterance clustering}. Each themed utterance is embedded with SentenceBERT ({\tt all-mpnet-base-v2} model is used, \citealt{reimers-2019-sentence-bert}), then the embeddings are clustered using the K-means algorithm \cite{Jin2010} with 10 clusters by default and the {\tt k-means++} initializer \cite{10.5555/1283383.1283494}.
 \item \textbf{Theme cluster adjustment to user preferences}. We apply a naïve algorithm that re-assigns cluster labels for every utterance id containing in the should-link / cannot-link sets. For every $<utt_i, utt_j>$ pair in the should-link set, if they are assigned to different clusters, $utt_j$ is re-assigned to $utt_i$'s cluster. In turn, for every $<utt_m, utt_n>$ in the cannot-link set, $utt_n$ is re-assigned to the cluster with the second closest centroid to it. Evidently, the baseline cluster adjustment algorithm doesn't have any generalization outside of the given preference sets.
 \item \textbf{Theme label generation}. We used an LLM with the prompt as in Appendix \ref{sec:apx-theme-labeling-guideline}~--- the default model used in the baseline implementation is {\tt Mistral-7B-Instruct-v0.3} \cite{jiang2023mistral7b}. No limitation on the number of in-context utterances was set.
\end{enumerate}

%% file: eval.tex
\begin{table*}[!t]
\centering
\small
\caption{Automatic Evaluation~--- Theme Label Metrics. Here and below, results in \textbf{bold} are the best, \underline{underlined} are those above baseline.}
\label{tab:auto-eval-label}
\begin{tabular}{lrrrrrrrrrr}
\toprule
\textbf{Team ID} & \textbf{R-1} & \textbf{R-2} & \textbf{R-L} & \textbf{Cos sim} & \textbf{BERT P} & \textbf{BERT R} & \textbf{BERT F1} & \textbf{LLM s1} & \textbf{LLM s2} & \textbf{LLM avg} \\
\midrule
Team A & 32.70\% & 4.60\% & 29.82\% & 59.51\% & \underline{89.82\%} & \underline{91.20\%} & \underline{90.35\%} & \underline{46.01\%} & \underline{56.47\%} & \underline{51.24\%} \\
Team B & 5.03\% & 0.00\% & 5.03\% & 37.08\% & 85.22\% & 88.02\% & 86.53\% & 12.03\% & 0.13\% & 6.08\% \\
Team C & \textbf{45.22\%} & 23.81\% & \textbf{45.10\%} & \textbf{69.91\%} & \textbf{95.02\%} & \textbf{94.69\%} & \textbf{94.71\%} & \textbf{100.00\%} & \textbf{99.48\%} & \textbf{99.74\%} \\
Team D & 34.57\% & 21.31\% & 34.27\% & 55.93\% & \underline{92.52\%} & \underline{91.48\%} & \underline{91.91\%} & \underline{80.39\%} & \underline{76.60\%} & \underline{78.50\%} \\
Team E & 42.28\% & 16.50\% & 41.22\% & \underline{62.48\%} & \underline{93.85\%} & \underline{92.84\%} & \underline{93.27\%} & \underline{93.46\%} & \underline{95.69\%} & \underline{94.58\%} \\
Team F & 23.10\% & 0.79\% & 21.14\% & 46.02\% & 85.67\% & 89.29\% & 87.19\% & 4.05\% & 3.53\% & 3.79\% \\\midrule
Baseline & 43.74\% & \textbf{24.56}\% & 42.87\% & 59.68\% & 89.25\% & 89.87\% & 89.52\% & 20.39\% & 39.48\% & 29.93\% \\
BL-prefs & 29.27\% & 4.21\% & 24.69\% & 48.79\% & 85.31\% & 87.77\% & 86.44\% & 12.81\% & 18.43\% & 15.62\%\\
\bottomrule
\end{tabular}
\end{table*}

Theme assignment that is the result of our task's solution can be assessed from two perspectives: from the controlled clustering perspective and from the theme label generation perspective~--- our evaluation metrics reflect these two perspectives.

\subsection{Automatic evaluation}
\label{sec:auto-eval}

Automatic evaluation metrics are mainly used for the development purposes and were provided to the participants as part of the starter code.

\subsubsection{Clustering metrics}
\label{sec:auto-eval-clustering}

\begin{itemize}
    \item \textbf{NMI} score \cite{JMLR:v11:vinh10a}~--- \textit{Normalized Mutual Information} is a function that measures the agreement of the two cluster assignments, reference and predicted, ignoring permutations. Normalization is performed over the mean of the entropies of the two assignments 
    \item \textbf{ACC} score \cite{6976982} evaluates the optimal alignment between the reference cluster assignment and the predicted one, with the alignment obtained using the Hungarian algorithm.
\end{itemize}

\subsubsection{Label generation metrics}
\label{sec:auto-eval-label}
We evaluate the predicted labels for theme clusters in two general ways: 1) similarity to the reference labels, 2) adherence to the theme label guideline.

Similarity of a predicted label to the references is calculated in the following way:

$$
Score_{i}(Y_i, \hat{y_i}) = \max_j{sim(Y_{i,j}, \hat{y_i})}
$$

where $Y_i$ are the reference labels for the $i$-th utterance (we provide two labels with a more specific and a more vague wording, respectively), $y_i$ is the predicted label for the same utterance and $sim$ is one of the similarity functions listed below.

\begin{itemize}
    \item \textbf{Cosine similarity}~--- the semantic similarity measure over SentenceBERT embeddings ({\tt all-mpnet-base-v2} model is used, \citealt{reimers-2019-sentence-bert}) of the reference and predicted labels,
    \item \textbf{ROUGE} score \cite{lin-2004-rouge}~--- an token-level N-gram overlap metric useful for comparing short and concise word sequences,
    \item \textbf{BERTScore} \cite{DBLP:conf/iclr/ZhangKWWA20} combines the agility of embedding-based similarity and the interpretability of token-level overlap. The model tokenizes each utterance and generates a contextual embedding for each token. Then, a cosine similarity $sim_{i,j}$ is calculated between $i$-th token of the reference and $j$-th token of the prediction. We report BERTScore \textbf{Precision} (for each token in the prediction, finding the reference token with the highest similarity), \textbf{Recall} (for each token in the reference, finding the prediction token with the highest similarity) and \textbf{F1 score}.
\end{itemize}

Adherence to the guideline is evaluated with an LLM-as-a-Judge prompted with a version of the guideline attached in the Appendix \ref{sec:apx-theme-labeling-guideline} (it was provided for the participants). For the usage with the LLM, it was split into three sections spanning structural and functional criteria, i.e. how good the label is as a linguistic expression and how good it is as an analytical tool, respectively. For the sake of preventing evaluation metric hacking, we shared a different / condensed version of the guideline to the participants and kept the full version held out.  For evaluation during the development phase, our provided code used a self-hosted solution with {\tt vicuna-13b-v1.5} \cite{10.5555/3666122.3668142} as the default LLMaaJ backbone. In the automatic evaluation of the final submissions, we used Claude 3.5 Sonnet \cite{anthropic_claude_3_5_sonnet}. Our repository contains both the public version of the label style evaluation prompt (3 condensed sections optimized for usage with public self-hosted LLMs) and its held out version (2 expanded sections optimized for usage with Claude, uploaded after the end of the competition).

\subsection{Human Evaluation}
\label{sec:human-eval}

All submissions underwent expert human evaluation in order to verify automated evaluation results and to expand the automated evaluation methodology to more precisely assess each solution's performance. The evaluation dimensions were divided into two broad categories covering formal and functional criteria, and each of these areas had additional subdimensions to be rated by evaluators in a binary fashion (\emph{pass}/\emph{fail}) using criteria distributed into into two broad categories: Structural/Functional. The structural criteria were based on the theme labeling guidelines provided to participants.

\begin{itemize}[leftmargin=0pt, label={}]
    \item \textbf{Structural Criteria} (Theme Label as a Linguistic Expression): Conciseness \& Word Choice, Grammatical Structure
    \item \textbf{Functional Criteria}\footnote{All functional criteria dimensions were evaluated at the level of the utterance except for \emph{Thematic Distinctiveness}, which was evaluated for each cluster label.} (Theme Label as an Analytical Tool): Semantic Relevance, Analytical Utility, Granularity, Actionability, Domain Relevance, Thematic Distinctiveness.
\end{itemize}

The guidelines for each of these dimensions, along with the positive and negative examples provided to evaluators (with reasoning), are laid out in Appendix \ref{sec:apx-human-eval-guidelines}. The theme labeling guidelines, upon which the structural criteria were based, are defined in Appendix \ref{sec:apx-theme-labeling-guideline}. The annotation task was completed in a single-pass way by two members of the track organizing team.

%% file: results.tex
\begin{table*}
\centering
\small
\caption{Human Evaluation~--- Per-utterance Functional Metrics}
\label{tab:human-eval-per-utterance}
\begin{tabular}{lrrrrr}
\toprule
\textbf{Team ID} & \textbf{Semantic Relevance} & \textbf{Analytical Utility} & \textbf{Granularity} & \textbf{Actionability} & \textbf{Domain Relevance} \\
\midrule
Team A & 77.25\% & 63.66\% & 22.75\% & 56.21\% & 79.74\% \\
Team B & 64.97\% & 12.94\% & 0.00\% & 4.05\% & 97.78\% \\
Team C & \textbf{89.67\%} & \textbf{82.75\%} & 47.84\% & \textbf{74.77\%} & \textbf{98.82\%} \\
Team D & 68.76\% & 63.66\% & 26.41\% & 60.26\% & 94.25\% \\
Team E & 86.27\% & 54.64\% & 22.48\% & 54.51\% & 91.11\% \\
Team F & 45.23\% & 41.57\% & 7.71\% & 41.57\% & 67.45\% \\\midrule
Baseline & 86.61\% & 66.84\% & \textbf{47.98\%} & 66.84\% & 89.6\% \\
BL-prefs & 88.76\% & 42.09\% & 20.00\% & 42.09\% & 83.92\% \\
\bottomrule
\end{tabular}
\end{table*}

\begin{table*}
\centering
\small
\caption{Human Evaluation~--- Per-cluster Metrics}
\label{tab:human-eval-per-cluster}
\begin{tabular}{lrrr}
\toprule
& \multicolumn{2}{c}{\textbf{Structural}} & \multicolumn{1}{c}{\textbf{Functional}} \\
\cmidrule(lr){2-3} \cmidrule(lr){4-4}
\textbf{Team ID} & \textbf{Conciseness} & \textbf{Grammatical Structure} & \textbf{Thematic Distinctiveness} \\
\midrule
Team A & \underline{83.33\%} & \textbf{100.00\%} & 75.76\% \\
Team B & \textbf{100.00\%} & \underline{33.33\%} & 0.00\% \\
Team C & \textbf{100.00\%} & \textbf{100.00\%} & \textbf{91.11\%} \\
Team D & \underline{91.67\%} & \underline{66.67\%} & 90.91\% \\
Team E & \underline{93.65\%} & \underline{93.65\%} & 78.34\% \\
Team F & \underline{95.00\%} & \textbf{100.00\%} & 72.63\% \\\midrule
Baseline & 80.00\% & 30.00\% & \textbf{91.11\%}\\
BL-prefs & 80.00\% & 20.00\% & 66.67\% \\
\bottomrule
\end{tabular}
\end{table*}

\begin{table}[H]
\centering
\small
\caption{Automatic Evaluation~--- Clustering Metrics}
\label{tab:auto-eval-clustering}
\begin{tabular}{lrr}
\toprule
\textbf{Team ID} & \textbf{ACC} & \textbf{NMI} \\
\midrule
Team A & 48.37\% & 42.02\%  \\
Team B & 17.91\% & 1.97\%  \\
Team C & \textbf{67.97\%} & \textbf{70.39\%} \\
Team D & 51.76\% & 47.71\% \\
Team E & 35.82\% & 47.73\% \\
Team F & 26.67\% & 9.06\% \\\midrule
Baseline & 53.2\% & 50.59\% \\
BL-prefs & 47.97\% & 45.39\% \\
\bottomrule
\end{tabular}
\end{table}

We received submissions from 6 participant teams. During the development, the teams were free to use the provided public data across 3 domains for creating their own train / development setups and testing e.g. out-of-domain generalization of their approaches. The test domain was made public during the last week of the competition. When submitting the inference results via an online form, the participant teams were asked to provide a brief info about their approaches. Below are the questions and the summaries of the submitted answers:
\begin{itemize}[leftmargin=0pt, label={}]
  \item \textbf{What LLM type did you use?} (\textit{Open-source~--- self-hosted / Proprietary via API / No LLM / Other})
  
  Teams A, C and F used a proprietary API; teams B, D and E used an open-source self-hosted LLM.
  \item \textbf{How large of an LLM did you use?} \textit{(<30B / 30---100B / >100B / Unknown (proprietary API) / No LLM / Other})
  
  Team A, C and F's model size is unknown; teams B, D and E used a model with <30B parameters.
  
  \item \textbf{Did you use any conversational information (previous / past context of the utterance)? Please specify if yes}

  Team C used the context window of 5 turns; Team E used conversational context within the predicted topic segment.
  
  \item \textbf{What clustering algorithm did you use?}

  Team A used HDBScan \cite{10.1007/978-3-642-37456-2_14}; Teams B and D used K-Means \cite{Jin2010}; Team C used ClusterLLM \cite{zhang-etal-2023-clusterllm}; Team F experimented with K-Means, DBSCAN and HDBSCAN; Team E used Spectral Clustering \cite{868688}.
  
  \item \textbf{What text embedding model did you use?}

  Teams A and C used Instructor model \cite{su-etal-2023-one}; Teams B, D and E used SentenceBERT as per the baseline.
  
  \item \textbf{Did you use an embedding dimensionality reduction technique? (Please specify which one if yes)}
  
  Teams A and E used UMAP \cite{2018arXivUMAP}.
  
  \item \textbf{Did you use a data augmentation technique (please specify what kind)?}

  Team A used Speech Acts as a data augmentation; Team B used SimCSE \cite{gao2021simcse}; Team E used contrastive learning to augment the limited unlabeled data.
  
  \item \textbf{How did you use the should-link / cannot-link pairs?}

  Teams A, B and D used the baseline approach. Team C used an LLM to re-assign the clusters for all the utterances from the should-link pairs. For the cannot-link pairs, the LLM was used to identify the utterance of the pair not belonging to the cluster, and then to make the re-assignment. Team E trained a reward model from the should-link and cannot-link pairs that was later incorporated into the clustering algorithm to impose soft constraints.
  
  \item \textbf{Did you use the theme label styleguide~--- if yes, how?}

  Team C used General Schema to extract verbs and nouns for each utterance in the cluster, then using those, they generated theme labels. Theme D instructed the labeling LLM to generate Verb-Object pairs. Teams E and F used the provided styleguide itself. Team F added it directly into the labeling LLM's prompt, and Team E modified and simplified it first.
  
  \item \textbf{Short (1-2 paragraph) description of your approach}
  
  \textbf{Team A} proposed a cluster-then-label framework for thematic clustering of utterances. First, they compute utterance embeddings using either Sentence Transformers, InBedder, or Instructor models depending on the embedding type. they then apply clustering (KMeans or HDBSCAN with UMAP-based dimensionality reduction) to group thematically similar utterances. Clustering is refined using manually provided should-link and cannot-link preference pairs, ensuring better alignment with human notions of similarity. After clustering, each cluster is labeled automatically by prompting an LLM (ChatGPT or Gemini Flash) with a batch of utterances, extracting a theme label and brief explanation. The resulting predicted labels are assigned back to utterances, forming the final output for evaluation. This approach leverages both unsupervised structure discovery and lightweight LLM-based supervision for scalable and interpretable theme labeling.
  
  \textbf{Team B} used SCCL \cite{zhang-etal-2021-supporting} and applied SimCSE for data augmentation. After training the SCCL, they clustered the utterances with K-Means. They performed hyperparameter search for the number of clusters based on the Silhouette score and set it to 7. User preference data was not used.
  
  \textbf{Team C} first extracted keyphrases from conversations using an LLM. They also determined the numbers of clusters based on the Silhouette coefficient. Clustering was performed using ClusterLLM, and the embedder was fine-tuned on the clustered utterances. Subsequently, among the two candidates with the highest preference pair accuracy, the candidate with the greater number of clusters was selected as the final model. Utterances were then adjusted according to the preference pairs. Finally, for the clustered utterances, a general schema was extracted in terms of verbs and nouns, and based on both the schema and the utterance content, the final theme labels were generated.
  
  \textbf{Team D} explored two approaches. The first one involved designing a prompting strategy to generate concise labels in a Verb–Object format (e.g., \textit{``update address''}, \textit{``book flight''}), allowing for more structured and comparable cluster representations. The second approach used {\tt LLaMA-3.1-8B-Instruct} to evaluate whether two utterances (with dialog history) belonged to the same cluster, based on their distance from the cluster center. The second method showed limited performance, and they submitted results using the first one, with a more robust prompting-based labeling strategy.
  
  \textbf{Team E} propose PrefSegGen, a preference-aware topic segmentation and generation framework that addresses low-resource conversational theme understanding by integrating topical-structured context modeling with user-preference-aligned theme generation. First, they introduce a novel two-stage self-supervised contrastive learning topic segmentation framework to obtain the topic segment to which the target utterance belongs under low-resource conditions. It initially leverages the unlabeled dialogues to pretrain topic encoders ({\tt bert-base-uncased} \& {\tt sup-simcse-bert-base-uncased}) on coherence and similarity patterns, followed by supervised fine-tuning with minimal labeled data to enhance segmentation precision. Subsequently, they incorporate a reward-guided clustering mechanism to guarantee that the generated themes are both contextually grounded and preference-aligned. A reward model, trained on should-link and cannot-link pairs, dynamically assigns linkage weights that reflect semantic proximity in line with user expectations. These weights guide spectral clustering after UMAP-based embedding reduction. Crucially, for each target utterance, they utilize its segmented topical context as input when prompting {\tt LLaMA3-8B-Instruct}, coupled with the official style guide, to generate hierarchical theme labels. An ensemble refinement process further enhances topic consistency by filtering low-frequency labels, yielding final outputs that are structurally coherent, context-aware, and tailored to user preferences.
  
  \textbf{Team F} employed a large language model (LLM) to annotate utterances based on preference signals, and subsequently attempted to merge clusters according to the LLM-based annotation.
\end{itemize}

Our evaluation results reveal that \textbf{Team C}'s approach achieved the highest accuracy across the board on both human and automatic metrics. It was tied with \textbf{Team B} on Label Conciseness, and with \textbf{Teams A} and \textbf{F} on Grammatical Structure. Although only \textbf{Team C}'s approach achieves \textbf{100\%} on both, signifying that its label generation works in full accordance with the styleguide. \textbf{Team C} was also the only one to surpass the baseline on automatic clustering metrics. \textbf{Team E} achieved the second best overall performance in both automatic and human evaluation and \textbf{Team D} placed third.

It is noteworthy that for all the three winning places, the ranking induced by the automatic metrics matched that by the humans~--- indicating that 1) automatic similarity metrics are applicable for short text, and 2) automatic evaluation of higher-level concepts like our label guideline is sufficiently accurate with frontier LLMs-as-Judges.

%% file: conclusions.tex
In this paper, we introduced \textit{Theme Detection} as a critical task in conversational analytics, and the associated \textit{Controllable Conversational Theme Detection} competition track at Dialog System Technology Challenge (DSTC) 12~--- where joint theme clustering and cluster label generation was further combined with the custom theme cluster granularity controllable via the provided preference data.

We gave an overview of the competition setup including the problem, the benchmark dataset and the details of evaluation, both automatic and human. We presented the participant team's submissions and gave an analysis of the insights from those.

We hope that this new problem, together with the dataset and the insights obtained from the competition will foster further research and advancements in Conversational AI.

%% file: apx_theme_labeling_prompt.tex
\begin{lstlisting}
<task>
You are an expert call center assistant.
You will be given a set of utterances in
<utterances> </utterances> tags, each
one on a new line.
The utterances are part of call center
conversations between the customer and
the support agent.
Your task is to generate a short label
describing the theme of all the given
utterances. The theme label should be
under 5 words and describe the desired
customer's action in the call.

<guidance>
Output your response in the following
way.

<theme_label_explanation>
Your short step-by-step explanation
behind the theme
</theme_label_explanation>

<theme_label>
your theme label
</theme_label>

</guidance>
</task>

H:
<utterances>
{utterances}
</utterances>
\end{lstlisting}

%% file: apx_theme_labeling_guideline.tex
An acceptable theme label is structurally and semantically well-formed according to the rules outlined in this appendix. \emph{Structurally well-formed} means that the words and their arrangement in the theme label are acceptable. \emph{Semantically well-formed} means that the meaning and usability of the theme label are acceptable.

\subsection{Theme labels exclude unneeded and undesirable words.}

Theme labels should be concise (2–5 words long). They should only include essential words (see \ref{sec:label-gl-word-types} and \ref{sec:label-gl-words-exx} below). Essential words will primarily include content (open-class) words. Function (closed-class) words should be excluded. Prepositions may be included as needed but should be avoided when there is a synonymous alternative label without a preposition.

Theme labels should also exclude context-sensitive words like pronouns (\emph{him}, \emph{her}, \emph{them}, \emph{it}, \emph{us}, etc.) and demonstratives (\emph{this}, \emph{that}, \emph{those}, etc.).

\subsection{Word types}\label{sec:label-gl-word-types}

\begin{itemize}
    \item Content/open-class words:
    \begin{itemize}
        \item nouns (\emph{items}, \emph{insurance}, \emph{information}, \emph{order}, etc.)
        \item main verbs (\emph{check}, \emph{inquire}, \emph{add}, \emph{explore}, etc.)
        \item adjectives (\emph{new} patient, \emph{missing} item, etc.)
        \item other modifying words (\emph{shipping} information, \emph{product} options, etc.)
    \end{itemize}
    \item Function/closed-class words:
    \begin{itemize}
        \item articles/determiners (\emph{the}, \emph{a}, etc.)
        \item auxiliary verbs (\emph{have} or \emph{be}, as in \emph{I have eaten} or \emph{I am eating})
        \item copulas 
        \item negation (\emph{not} or \emph{-n't}, as in \emph{not on time} or \emph{didn't arrive})
        \item conjunctions (\emph{and}, \emph{or}, \emph{but}, etc.)
        \item complementizers (clause-embedding uses of \emph{that}, \emph{for}, \emph{if}, \emph{whether}, \emph{because}, etc.)
        \item modals (\emph{can}, \emph{could}, \emph{will}, \emph{would}, \emph{may}, \emph{might}, \emph{must}, \emph{shall})
        \item question words (\emph{who}, \emph{what}, \emph{where}, \emph{when}, \emph{how}, \emph{why}) 
    \end{itemize}
    \item Context-sensitive words:
    \begin{itemize}
        \item pronouns (\emph{she}, \emph{he}, \emph{they}, \emph{it}, \emph{her}, \emph{his}, etc.)
        \item demonstratives (\emph{this}, \emph{these}, \emph{that}, \emph{those}, etc.)
        \item temporal adverbs (\emph{yesterday}, \emph{tomorrow}, \emph{next week}, etc.)
        \item other context-sensitive language
        \begin{itemize}
            \item \emph{one}, as in \emph{I’m looking for a nearby branch. Can you find \emph{one}?}
            \item deleted nouns (noun ellipsis), as in \emph{I found his order, but not yours \_\_}.
        \end{itemize} 
    \end{itemize}
\end{itemize}

\subsubsection{Examples}\label{sec:label-gl-words-exx}

For a theme covering order tracking:
\begin{itemize}
    \item \textbf{\textcolor{green}{Good}}: track order
    \item \textbf{\textcolor{green}{Good}}: track shipment
    \item \textbf{\textcolor{red}{Bad}}: track an order (includes an article)
    \item \textbf{\textcolor{red}{Bad}}: track their order (includes a pronoun)
    \end{itemize}

For a theme covering finding the nearest branch of a chain:
\begin{itemize}
    \item \textbf{\textcolor{green}{Good}}: find nearest branch
    \item \textbf{\textcolor{green}{Good}}: find closest branch
    \item \textbf{\textcolor{red}{Bad}}: find nearest one (includes context-sensitive \emph{one})
    \item \textbf{\textcolor{red}{Bad}}: check if there’s a nearby branch (includes a complementizer \emph{if}; includes a form of \emph{be})
\end{itemize}

\subsection{Theme labels are verb phrases that classify events.}

A verb phrase begins with a verb and may include arguments or modifiers of the verb (such as a direct object). The verb should be in its citation form, lacking any complex morphology such as tense or agreement suffixes. The citation form of a verb is what would normally follow the infinitive \emph{to}, such as \emph{sign up} in \emph{I’d like to \emph{sign up}}. Theme labels should not be other phrase types, such as noun phrases.

The verb phrase should describe a class of events. Events are things that can be said to \textbf{happen}, unlike states (e.g. \emph{learn} [event] vs. \emph{know} [state]), entities (e.g. \emph{redeem} [event] vs. \emph{redemption} [entity]), properties (e.g. \emph{complain} [event] vs. \emph{angry} [property]), and claims (\emph{report defect} [event] vs. \emph{product is defective} [claim]).

\subsubsection{Examples}

For a theme covering membership sign-ups:
\begin{itemize}
    \item \textbf{\textcolor{green}{Good}}: sign up for membership (verb phrase; describes a kind of \emph{signing up} event)
    \item \textbf{\textcolor{red}{Bad}}: signing up for membership (verb phrase, but verb is not in citation form)
    \item \textbf{\textcolor{red}{Bad}}: membership sign-up (noun phrase; describes a kind of entity)
    \item \textbf{\textcolor{red}{Bad}}: memberships (noun phrase; describes a kind of entity)
\end{itemize}
For a theme covering requests to check in early at a hotel:
\begin{itemize}
    \item \textbf{\textcolor{green}{Good}}: request early check-in (verb phrase; describes a kind of requesting event)
    \item \textbf{\textcolor{red}{Bad}}: requested early check-in (verb phrase, but verb is not in citation form)
    \item \textbf{\textcolor{red}{Bad}}: request for early check-in (noun phrase; describes a kind of entity)
    \item \textbf{\textcolor{red}{Bad}}: customer wants early check-in (this is a claim)
\end{itemize}
For a theme covering reporting a defective product:
\begin{itemize}
    \item \textbf{\textcolor{green}{Good}}: report defective product (verb phrase; describes events)
    \item \textbf{\textcolor{red}{Bad}}: reporting defective product (verb phrase, but verb is not in citation form)
    \item \textbf{\textcolor{red}{Bad}}: believe product is defective (verb phrase, but describes a state rather than an event)
    \item \textbf{\textcolor{red}{Bad}}: defective product (noun phrase; describes a kind of entity)
\end{itemize}

\subsection{Theme labels are informative and actionable yet sufficiently general.}

Theme labels should be informative enough to substantially narrow down the set of possible customer issue resolution steps (the steps to resolve the problem/need that drove the customer to make contact). For example, \emph{check balance} is probably associated with a standard procedure for checking the balance of a range of customer account types, but \emph{perform check} is so broad that it could be associated with an extremely diverse group of issue resolutions. Non-actionable theme labels may be excessively vague or uninformative, and hence not very useful.

\subsubsection{Examples}

For a theme covering appointment-scheduling themes:
\begin{itemize}
    \item \textbf{\textcolor{green}{Good}}: schedule appointments
    \item \textbf{\textcolor{red}{Bad}}: ask about appointments (probably too general)
    \item \textbf{\textcolor{red}{Bad}}: schedule appointment for next week (too specific)
    \item \textbf{\textcolor{red}{Bad}}: schedule appointment for elderly parent (too specific)
\end{itemize}
For a theme covering adding a recognized user to an existing account or policy:
\begin{itemize}
    \item \textbf{\textcolor{green}{Good}}: add user
    \item \textbf{\textcolor{red}{Bad}}: add one (too general)
    \item \textbf{\textcolor{red}{Bad}}: add oldest child (too specific)
\end{itemize}
For a theme covering user password issues:
\begin{itemize}
    \item \textbf{\textcolor{green}{Good}}: reset password
    \item \textbf{\textcolor{green}{Good}}: troubleshoot password
    \item \textbf{\textcolor{red}{Bad}}: secure account (too general)
    \item \textbf{\textcolor{red}{Bad}}: reset password again (too specific)
\end{itemize}
For a theme covering credit or debit card charge disputes:
\begin{itemize}
    \item \textbf{\textcolor{green}{Good}}: dispute charge
    \item \textbf{\textcolor{red}{Bad}}: complain about charge (too general)
    \item \textbf{\textcolor{red}{Bad}}: file card complaint (too general)
    \item \textbf{\textcolor{red}{Bad}}: dispute charge for defective blender (too specific)
\end{itemize}

%% file: apx_human_eval_guidelines.tex
\subsection{Structural Dimensions}

\subsubsection{Conciseness \& Word Choice}

\textbf{Options}: Pass (1) / Fail (0)

\noindent\textbf{Definition}: The following criteria are consolidated by the evaluator into one Pass/Fail rating for Conciseness \& Word Choice:

\begin{enumerate}
    \item \textbf{Label length}: Is the label concise, containing only 2--5 words?
    \begin{itemize}
        \item \textbf{\textcolor{green}{Pass}}: update billing address\newline\textbf{\textcolor{red}{Fail}}: update customer's residential billing address for future statements\newline\textbf{Rationale}: The good example uses 3 words, within the required 2-5 word range. The bad example uses 8, making it unnecessarily verbose when the core intent can be expressed more concisely.
        \item \textbf{\textcolor{green}{Pass}}: access account statement\newline\textbf{\textcolor{red}{Fail}}: statement\newline\textbf{Rationale}: The good example uses 3 words, adhering to the 2-5 word guideline. The bad example uses only one word, which lacks sufficient specificity to be useful as a theme label.
    \end{itemize}
    \item \textbf{Function word exclusion}: Does the label exclude unnecessary function words (articles, auxiliary verbs, etc.)?
    \begin{itemize}
        \item \textbf{\textcolor{green}{Pass}}: add dependent coverage\newline\textbf{\textcolor{red}{Fail}}: add the dependent to coverage\newline\textbf{Rationale}: The good example correctly excludes function words like articles (``the''), focusing only on essential content words. The bad example unnecessarily includes ``the'', which should be excluded according to guidelines.
        \item \textbf{\textcolor{green}{Pass}}: troubleshoot internet connection\newline\textbf{\textcolor{red}{Fail}}: troubleshoot why internet is not working\newline\textbf{Rationale}: The good example properly excludes function words, while the bad example improperly includes function words ``why,'' ``is,'' and ``not'' which should be excluded for conciseness.
    \end{itemize}
    \item \textbf{Avoidance of context sensitivity}: Does the label exclude context-dependent words (pronouns, demonstratives, temporal adverbs, etc.)?
    \begin{itemize}
        \item \textbf{\textcolor{green}{Pass}}: return defective product\newline\textbf{\textcolor{red}{Fail}}: return this item\newline\textbf{Rationale}: The good example avoids context-sensitive words like ``this'' and uses the general term ``product'' that can apply across contexts. The bad example includes the context-sensitive demonstrative ``this,'' which requires a specific context to understand its meaning.
        \item \textbf{\textcolor{green}{Pass}}: reschedule appointment\newline\textbf{\textcolor{red}{Fail}}: reschedule it for tomorrow\newline\textbf{Rationale}: The good example uses general terminology applicable to any appointment, while the bad example includes both the pronoun ``it'' and the temporal adverb ``tomorrow,'' both of which are dependent on conversation context for their meaning.
    \end{itemize}
    \item \textbf{Preposition usage}: Are prepositions included only when necessary?
    \begin{itemize}
        \item \textbf{\textcolor{green}{Pass}}: transfer funds\newline\textbf{\textcolor{red}{Fail}}: transfer from account\newline\textbf{Rationale}: The good example avoids unnecessary prepositions by using a concise verb-object structure. The bad example unnecessarily includes the preposition ``from'' when the more concise alternative without the preposition works just as well.
        \item \textbf{\textcolor{green}{Pass}}: join rewards program\newline\textbf{\textcolor{red}{Fail}}: sign up for rewards program\newline\textbf{Rationale}: The good example avoids prepositions entirely, while the bad example unnecessarily includes the preposition ``for'' when alternatives without prepositions are available and equally clear.
    \end{itemize}
\end{enumerate}

\subsubsection{Grammatical Structure}

\textbf{Options}: Pass (1) / Fail (0)

\noindent\textbf{Definition}: The following criteria are consolidated by the evaluator into one Pass/Fail rating for Grammatical Structure:

\begin{enumerate}
    \item \textbf{Verb phrase structure}: Is the label a verb phrase?
    \begin{itemize}
        \item \textbf{\textcolor{green}{Pass}}: cancel flight\newline\textbf{\textcolor{red}{Fail}}: flight cancellation\newline\textbf{Rationale}: The good example correctly follows the verb phrase requirement by starting with a verb (``cancel'') followed by a noun (``flight''). The bad example uses a noun phrase (``flight cancellation'') instead.
        \item \textbf{\textcolor{green}{Pass}}: redeem rewards\newline\textbf{\textcolor{red}{Fail}}: rewards redemption process\newline\textbf{Rationale}: The good example uses a verb phrase beginning with the verb ``redeem''. The bad example fails by using a noun phrase with ``redemption'' as the head noun rather than using a verb form.
    \end{itemize}
    \item \textbf{Citation form}: Does the verb appear in its citation form (without tense or agreement morphology)?
    \begin{itemize}
        \item \textbf{\textcolor{green}{Pass}}: change delivery address\newline\textbf{\textcolor{red}{Fail}}: changing delivery address\newline\textbf{Rationale}: The good example correctly uses the citation form of the verb ``change'' without any tense or agreement morphology. The bad example fails by using the -ing form ``changing'' rather than the required base form.
        \item \textbf{\textcolor{green}{Pass}}: cancel subscription\newline\textbf{\textcolor{red}{Fail}}: canceled subscription\newline\textbf{Rationale}: The good example properly uses the citation form of the verb ``cancel'' without inflectional endings. The bad example incorrectly uses the past tense form ``cancelled'' instead of the citation form.
    \end{itemize}
    \item \textbf{Event classification}: Does the verb phrase describe a class of events, rather than states, entities, properties, or claims?
    \begin{itemize}
        \item \textbf{\textcolor{green}{Pass}}: verify warranty coverage\newline\textbf{\textcolor{red}{Fail}}: warranty coverage\newline\textbf{Rationale}: The good example describes an event (the act of verifying) rather than an entity. The bad example describes an entity (the warranty coverage itself) rather than an event, violating the requirement that theme labels classify events. Note: The bad example would also be ruled out by the verb phrase requirement.
        \item \textbf{\textcolor{green}{Pass}}: express dissatisfaction\newline\textbf{\textcolor{red}{Fail}}: customer is dissatisfied\newline\textbf{\textcolor{red}{Fail}}: is dissatisfied\newline\textbf{Rationale}: The good example describes an event (the act of expressing) rather than a state. The first bad example is structured as a claim about the customer, rather than describing en event. The second bad example is a verb phrase but describes the wrong kind of situation: a state, rather than an event.
        \item \textbf{\textcolor{green}{Pass}}: complain about faulty product (event)\newline\textbf{\textcolor{red}{Fail}}: angry about faulty product (property)\newline\textbf{Rationale}: The good example describes an event (the act of complaining) rather than a property. The bad example describes a property or attribute of the customer, rather than an event describing the customer's intent.
    \end{itemize}
\end{enumerate}

\subsection{Functional Dimensions}

\subsubsection{Semantic Relevance}

\textbf{Options}: Pass (1) / Fail (0)

\noindent\textbf{Definition}: Does the label accurately capture the core intent/topic of the utterance it represents? Theme labels are expected to provide a gist of the dialogue from the customer's inquiry perspective.

\begin{itemize}
    \item \textbf{\textcolor{green}{Pass}}: request card security support (For customer utterance: ``I received a notification that my credit card might have been compromised. I need to know what steps I should take.'')\newline\textbf{Rationale}: This theme label demonstrates good semantic relevance by accurately capturing the core intent of the customer's inquiry—addressing a potential security issue—rather than focusing on peripheral aspects like the notification itself.
    \item \textbf{\textcolor{red}{Fail}}: express frustration (For customer utterance: ``I've been on hold for 45 minutes trying to get help with activating my new debit card. This is ridiculous!'')\newline\textbf{Rationale}: This theme label fails the semantic relevance test because it focuses on the customer's emotional state rather than their actual intent, which is to activate their debit card. The frustration is secondary to the core purpose of the contact.
    \item \textbf{\textcolor{green}{Pass}}: book accommodation\newline\textbf{\textcolor{red}{Fail}}: inquire about Chicago\newline\textbf{Rationale}: The good example correctly identifies the core intent (booking a hotel room), while the bad example misidentifies the intent as seeking information about Chicago when the location is just a detail/slot related to the booking request.
\end{itemize}

\subsubsection{Analytical Utility}

\textbf{Options}: Pass (1) / Fail (0)

\noindent\textbf{Definition}: Does the label provide meaningful categorization that could directly support a reviewer or analyst's workflow when reviewing conversation data? Themes, which should be ready for presentation to the user/analyst, are supposed to highlight the topics discussed in the conversation that are useful for categorizing and further analyzing them according to the nature of the conversation.

\begin{itemize}
    \item \textbf{\textcolor{green}{Pass}}: troubleshoot checkout error\newline\textbf{For customer utterance}: ``I'm getting error code E-503 when trying to complete my purchase on your website. I've tried three different browsers.''\newline\textbf{Rationale}: This theme label has good analytical utility because it categorizes the issue in a way that would allow analysts to, e.g., identify patterns in checkout problems, prioritize technical fixes, and track the frequency of specific error types.
    \item  \textbf{\textcolor{red}{Fail}}: customer contact\newline\textbf{For customer utterance}: ``I ordered a blue shirt in size medium last week, but you sent me a red one instead. I'd like to exchange it.''\newline\textbf{Rationale}: This theme label lacks analytical utility because it's too broad to provide meaningful categorization. It fails to identify the specific issue (there's an order fulfillment error) in a way that could help improve operations or track problem patterns.
    \item \textbf{\textcolor{green}{Pass}}: downgrade service plan\newline\textbf{\textcolor{red}{Fail}}: smart thermostat model TH8000 connection failure with iOS app version 3.2.1\newline\textbf{Rationale}: The good example provides useful categorization at the right level of detail for business analysis. The bad example is too specific with technical details that would fragment similar issues into tiny categories, making pattern identification difficult.
\end{itemize}

\subsubsection{Granularity}

\textbf{Options}: Pass (1) / Fail (0)

\noindent\textbf{Definition}: Does the label maintain appropriate specificity, as determined by its closeness to the provided gold labels? (Submission authors are expected to infer ideal granularity from the provided user preference data.)

\begin{itemize}
    \item  \textbf{\textcolor{green}{Pass}}: update payment information\newline\textbf{\textcolor{red}{Fail}}: manage account\newline\textbf{Rationale}: The good example demonstrates appropriate granularity by categorizing the issue at a level that's neither too broad nor too specific. The bad example is too broad, grouping potentially diverse issues that would benefit from more specific categorization.
    \item \textbf{\textcolor{green}{Pass}}: troubleshoot device connectivity\newline\textbf{\textcolor{red}{Fail}}: resolve Sony WH-1000XM4 headphones pairing failure with streaming app on Android 16 beta\newline\textbf{Rationale}: The good example shows appropriate granularity by categorizing at a level that groups similar technical problems. The bad example has excessive granularity, including specific device models and OS versions that would create overly-fragmented categories.
\end{itemize}

\subsubsection{Actionability}

\textbf{Options}: Pass (1) / Fail (0)

\noindent\textbf{Definition}: Does the label provide sufficient information to categorize customer issues for resolution? Theme labels should be informative enough to substantially narrow down the set of possible customer issue resolution steps.

\begin{itemize}
    \item \textbf{\textcolor{green}{Pass}}: dispute transaction\newline\textbf{\textcolor{red}{Fail}}: seek assistance\newline\textbf{Rationale}: The good example demonstrates good actionability by clearly identifying a specific process (transaction dispute) with established resolution procedures. The bad example is too vague to suggest any specific resolution path.
    \item \textbf{\textcolor{green}{Pass}}: trace missing shipment\newline\textbf{\textcolor{red}{Fail}}: discuss app features\newline\textbf{Rationale}: The good example shows good actionability by identifying a specific issue (shipment tracking problem) that points to clear resolution steps. The bad example has poor actionability because ``discuss'' doesn't point to a specific resolution-related action, and ``app features'' is too broad.
\end{itemize}

\subsubsection{Domain Relevance}

\textbf{Options}: Pass (1) / Fail (0)

\noindent\textbf{Definition}: Does the label reflect domain-specific terminology and concepts appropriate to the conversation context? Theme labels should reduce manual analysis by utilizing domain-relevant and context-relevant terminology.

\begin{itemize}
    \item \textbf{\textcolor{green}{Pass}}: verify coverage details\newline\textbf{For customer utterance}: ``I need to know if my insurance policy covers damage from a burst pipe in my basement.''\newline\textbf{Rationale}: This theme label demonstrates good domain relevance by using terminology (``verify coverage'') that's specific to the insurance industry and reflects how claims and policy questions are typically categorized in that domain.
    \item \textbf{\textcolor{green}{Pass}}: transfer prescription\newline\textbf{For customer utterance}: ``I want to transfer my prescription from my old pharmacy to your location. Can you help with that?''\newline\textbf{Rationale}: This theme label shows good domain relevance by using standard pharmacy industry terminology (``transfer prescription'') that accurately reflects how this process is categorized and handled within the healthcare/pharmacy domain.
    \item \textbf{\textcolor{red}{Fail}}: change money amount\newline\textbf{For customer utterance}: ``I need to increase my 401(k) contribution percentage starting with my next paycheck.''\newline\textbf{Rationale}: This theme label lacks domain relevance because it uses overly-generic terminology instead of financial industry-specific language. A more domain-relevant label would be ``adjust retirement contribution'' or ``modify investment allocation.''
    \item \textbf{\textcolor{red}{Fail}}: fix travel problem\newline\textbf{For customer utterance}: ``My flight was delayed and I missed my connection. I need to be rebooked on the next available flight.'')\newline\textbf{Rationale}: This theme label has poor domain relevance because it doesn't use airline industry terminology. A more domain-relevant label would be ``rebook missed connection'', ``accommodate disrupted itinerary'', etc.
\end{itemize}

\subsubsection{Thematic Distinctiveness}

\textbf{Options}: Pass (1) / Fail (0)

\noindent\textbf{Definition}: Does the label create a clear boundary that differentiates one theme from the other themes in the dataset? Theme labels should exhaustively cover all the examples AND be mutually exclusive.

\begin{itemize}
    \item \textbf{\textcolor{green}{Pass}}: report stolen card\newline\textbf{In this context}: Dataset already contains theme labels ``report lost card'' and ``report fraudulent transaction''\newline\textbf{For customer utterance}: ``Someone stole my wallet and I need to block my credit card immediately.''\newline\textbf{Rationale}: This theme label demonstrates good thematic distinctiveness by creating a clear boundary between related but distinct issues: lost cards (misplaced by owner), stolen cards (taken by someone else), and fraudulent transactions (unauthorized use).'
    \item \textbf{\textcolor{red}{Fail}}: inquire about refund\newline\textbf{In this context}: Dataset already contains theme label ``request refund''\newline\textbf{For customer utterance}: ``I returned my purchase last week but haven't seen the money back in my account yet.''\newline\textbf{Rationale}: This theme label fails the thematic distinctiveness test because it doesn't create a clear boundary between refund requests and refund status checks. The new utterance is about tracking a refund in progress, which should be a distinct category (e.g. ``check refund status''. Instead, this category could be compatible with utterances that are already covered by ``request refund''.
    \item \textbf{\textcolor{green}{Pass}}: change delivery location\newline\textbf{\textcolor{red}{Fail}}: reset account\newline\textbf{In this context}: Dataset already contains theme labels ``schedule delivery'', ``reschedule delivery'', ``reset password'', and ``update account information''\newline\textbf{Rationale}: The good example shows appropriate thematic distinctiveness by creating a clear boundary between different delivery modification types. The bad example blurs the boundary between password resets and other profile updates, creating confusion about categorization.
\end{itemize}

%% file: apx_io_examples.tex
Below is an input datapoint for a dialogue with one utterance marked as themed. For the train/dev domains, the theme labels will be available as in the example below. For the test domain, only the flag that an utterance is themed will be provided.

\begin{lstlisting}[language=Python]
{
  "conversation_id": "Banking_123",
  "turns": [
      {
        "speaker": "Agent",
        "utterance": "Thank you for calling Intellibank. This is Melanie. How can I help you?"
      },
      {
        "speaker": "Customer",
        "utterance": "Yeah, hey. This is John Smith. I've got a quick question."
      },
      {
        "speaker": "Agent",
        "utterance": "OK, John. What can I help you with?"
      },
      {
        "speaker": "Customer",
        "utterance": "Yeah I need to know what your ATM withdrawal limits are for the day.",
        "theme_label": "get daily withdrawal limit",
      },
      {
        "speaker": "Agent",
        "utterance": "Certainly. Our ATM withdrawal limit is on a per day basis and it is up to two hundred dollars."
      },
      {
        "speaker": "Customer",
        "utterance": "Oh perfect, perfect. Yeah, I think I'll just see if I can head down to the ATM now. Thank you."
      },
      {
        "speaker": "Agent",
        "utterance": "OK, thank you. You have a great day."
      },
      {
        "speaker": "Customer",
        "utterance": "You too."
      }
    ]
}

\end{lstlisting}

Below is an input datapoint with the example user preference on clustering granularity:

\begin{lstlisting}[language=Python]
{
  "utterance_a": {
    "utterance": "Yeah, so I need to change the account number thing that I put in whenever I go to the ATM."
    "conversation_id": "Banking_123",
    "turn_id": 4
  },
  "utterance_b": {
    "utterance": "OK. Excellent. Thank you Ms. Crystal. And while I got you on the phone I see it's been a little bit since you've authenticated your account here. Would you like to add a PIN number to your account for security reasons?"
    "conversation_id": "Banking_345",
    "turn_id": 10
  },
  "belong_to_same_theme": "yes"
}
\end{lstlisting}

%% file: acl2023.bbl
\begin{thebibliography}{26}
\expandafter\ifx\csname natexlab\endcsname\relax\def\natexlab#1{#1}\fi

\bibitem[{{Anthropic}(2024)}]{anthropic_claude_3_5_sonnet}
{Anthropic}. 2024.
\newblock Introducing {C}laude 3.5 {S}onnet.
\newblock \url{https://www.anthropic.com/news/claude-3-5-sonnet}.
\newblock Accessed: 2025-06-13.

\bibitem[{Arthur and Vassilvitskii(2007)}]{10.5555/1283383.1283494}
David Arthur and Sergei Vassilvitskii. 2007.
\newblock k-means++: the advantages of careful seeding.
\newblock In \emph{Proceedings of the Eighteenth Annual ACM-SIAM Symposium on Discrete Algorithms}, SODA '07, page 1027–1035, USA. Society for Industrial and Applied Mathematics.

\bibitem[{Basu et~al.(2004)Basu, Banerjee, and Mooney}]{DBLP:conf/sdm/BasuBM04}
Sugato Basu, Arindam Banerjee, and Raymond~J. Mooney. 2004.
\newblock \href {https://doi.org/10.1137/1.9781611972740.31} {Active semi-supervision for pairwise constrained clustering}.
\newblock In \emph{Proceedings of the Fourth {SIAM} International Conference on Data Mining, Lake Buena Vista, Florida, USA, April 22-24, 2004}, pages 333--344. {SIAM}.

\bibitem[{Campello et~al.(2013)Campello, Moulavi, and Sander}]{10.1007/978-3-642-37456-2_14}
Ricardo J. G.~B. Campello, Davoud Moulavi, and Joerg Sander. 2013.
\newblock Density-based clustering based on hierarchical density estimates.
\newblock In \emph{Advances in Knowledge Discovery and Data Mining}, pages 160--172, Berlin, Heidelberg. Springer Berlin Heidelberg.

\bibitem[{Chu et~al.(2023)Chu, Li, Liu, Gu, Liu, Ge, and Hu}]{chu-etal-2023-multi}
Caiyuan Chu, Ya~Li, Yifan Liu, Jia-Chen Gu, Quan Liu, Yongxin Ge, and Guoping Hu. 2023.
\newblock \href {https://aclanthology.org/2023.dstc-1.5} {Multi-stage coarse-to-fine contrastive learning for conversation intent induction}.
\newblock In \emph{Proceedings of The Eleventh Dialog System Technology Challenge}, pages 31--39, Prague, Czech Republic. Association for Computational Linguistics.

\bibitem[{Gao et~al.(2021)Gao, Yao, and Chen}]{gao2021simcse}
Tianyu Gao, Xingcheng Yao, and Danqi Chen. 2021.
\newblock {SimCSE}: Simple contrastive learning of sentence embeddings.
\newblock In \emph{Empirical Methods in Natural Language Processing (EMNLP)}.

\bibitem[{Gonz{\'{a}}lez{-}Almagro et~al.(2025)Gonz{\'{a}}lez{-}Almagro, Peralta, Poorter, Cano, and Garc{\'{\i}}a}]{DBLP:journals/air/GonzalezAlmagroPPCG25}
Germ{\'{a}}n Gonz{\'{a}}lez{-}Almagro, Daniel Peralta, Eli~De Poorter, Jos{\'{e}}~Ram{\'{o}}n Cano, and Salvador Garc{\'{\i}}a. 2025.
\newblock \href {https://doi.org/10.1007/S10462-024-11103-8} {Semi-supervised constrained clustering: an in-depth overview, ranked taxonomy and future research directions}.
\newblock \emph{Artif. Intell. Rev.}, 58(5):157.

\bibitem[{Gung et~al.(2023{\natexlab{a}})Gung, Moeng, Rose, Gupta, Zhang, and Mansour}]{gung-etal-2023-natcs}
James Gung, Emily Moeng, Wesley Rose, Arshit Gupta, Yi~Zhang, and Saab Mansour. 2023{\natexlab{a}}.
\newblock \href {https://doi.org/10.18653/v1/2023.findings-acl.613} {{N}at{CS}: Eliciting natural customer support dialogues}.
\newblock In \emph{Findings of the Association for Computational Linguistics: ACL 2023}, pages 9652--9677, Toronto, Canada. Association for Computational Linguistics.

\bibitem[{Gung et~al.(2023{\natexlab{b}})Gung, Shu, Moeng, Rose, Romeo, Gupta, Benajiba, Mansour, and Zhang}]{gung-etal-2023-intent}
James Gung, Raphael Shu, Emily Moeng, Wesley Rose, Salvatore Romeo, Arshit Gupta, Yassine Benajiba, Saab Mansour, and Yi~Zhang. 2023{\natexlab{b}}.
\newblock \href {https://aclanthology.org/2023.dstc-1.27} {Intent induction from conversations for task-oriented dialogue track at {DSTC} 11}.
\newblock In \emph{Proceedings of The Eleventh Dialog System Technology Challenge}, pages 242--259, Prague, Czech Republic. Association for Computational Linguistics.

\bibitem[{Huang et~al.(2014)Huang, Huang, Wang, and Wang}]{6976982}
Peihao Huang, Yan Huang, Wei Wang, and Liang Wang. 2014.
\newblock \href {https://doi.org/10.1109/ICPR.2014.272} {Deep embedding network for clustering}.
\newblock In \emph{2014 22nd International Conference on Pattern Recognition}, pages 1532--1537.

\bibitem[{Jiang et~al.(2023)Jiang, Sablayrolles, Mensch, Bamford, Chaplot, de~las Casas, Bressand, Lengyel, Lample, Saulnier, Lavaud, Lachaux, Stock, Scao, Lavril, Wang, Lacroix, and Sayed}]{jiang2023mistral7b}
Albert~Q. Jiang, Alexandre Sablayrolles, Arthur Mensch, Chris Bamford, Devendra~Singh Chaplot, Diego de~las Casas, Florian Bressand, Gianna Lengyel, Guillaume Lample, Lucile Saulnier, Lélio~Renard Lavaud, Marie-Anne Lachaux, Pierre Stock, Teven~Le Scao, Thibaut Lavril, Thomas Wang, Timothée Lacroix, and William~El Sayed. 2023.
\newblock \href {http://arxiv.org/abs/2310.06825} {Mistral 7b}.

\bibitem[{Jin and Han(2010)}]{Jin2010}
Xin Jin and Jiawei Han. 2010.
\newblock \href {https://doi.org/10.1007/978-0-387-30164-8_425} {\emph{K-Means Clustering}}, pages 563--564. Springer US, Boston, MA.

\bibitem[{Lin(2004)}]{lin-2004-rouge}
Chin-Yew Lin. 2004.
\newblock \href {https://aclanthology.org/W04-1013} {{ROUGE}: A package for automatic evaluation of summaries}.
\newblock In \emph{Text Summarization Branches Out}, pages 74--81, Barcelona, Spain. Association for Computational Linguistics.

\bibitem[{Manning et~al.(2008)Manning, Raghavan, and Schütze}]{Manning_Raghavan_Schütze_2008}
Christopher~D. Manning, Prabhakar Raghavan, and Hinrich Schütze. 2008.
\newblock \emph{Introduction to Information Retrieval}.
\newblock Cambridge University Press.

\bibitem[{{McInnes} et~al.(2018){McInnes}, {Healy}, and {Melville}}]{2018arXivUMAP}
L.~{McInnes}, J.~{Healy}, and J.~{Melville}. 2018.
\newblock \href {http://arxiv.org/abs/1802.03426} {{UMAP: Uniform Manifold Approximation and Projection for Dimension Reduction}}.
\newblock \emph{ArXiv e-prints}.

\bibitem[{Reimers and Gurevych(2019)}]{reimers-2019-sentence-bert}
Nils Reimers and Iryna Gurevych. 2019.
\newblock \href {https://arxiv.org/abs/1908.10084} {Sentence-bert: Sentence embeddings using siamese bert-networks}.
\newblock In \emph{Proceedings of the 2019 Conference on Empirical Methods in Natural Language Processing}. Association for Computational Linguistics.

\bibitem[{Shi and Malik(2000)}]{868688}
Jianbo Shi and J.~Malik. 2000.
\newblock \href {https://doi.org/10.1109/34.868688} {Normalized cuts and image segmentation}.
\newblock \emph{IEEE Transactions on Pattern Analysis and Machine Intelligence}, 22(8):888--905.

\bibitem[{Su et~al.(2023)Su, Shi, Kasai, Wang, Hu, Ostendorf, Yih, Smith, Zettlemoyer, and Yu}]{su-etal-2023-one}
Hongjin Su, Weijia Shi, Jungo Kasai, Yizhong Wang, Yushi Hu, Mari Ostendorf, Wen-tau Yih, Noah~A. Smith, Luke Zettlemoyer, and Tao Yu. 2023.
\newblock \href {https://doi.org/10.18653/v1/2023.findings-acl.71} {One embedder, any task: Instruction-finetuned text embeddings}.
\newblock In \emph{Findings of the Association for Computational Linguistics: ACL 2023}, pages 1102--1121, Toronto, Canada. Association for Computational Linguistics.

\bibitem[{Vinh et~al.(2010)Vinh, Epps, and Bailey}]{JMLR:v11:vinh10a}
Nguyen~Xuan Vinh, Julien Epps, and James Bailey. 2010.
\newblock \href {http://jmlr.org/papers/v11/vinh10a.html} {Information theoretic measures for clusterings comparison: Variants, properties, normalization and correction for chance}.
\newblock \emph{Journal of Machine Learning Research}, 11(95):2837--2854.

\bibitem[{Viswanathan et~al.(2024)Viswanathan, Gashteovski, Gashteovski, Lawrence, Wu, and Neubig}]{viswanathan-etal-2024-large}
Vijay Viswanathan, Kiril Gashteovski, Kiril Gashteovski, Carolin Lawrence, Tongshuang Wu, and Graham Neubig. 2024.
\newblock \href {https://doi.org/10.1162/tacl_a_00648} {Large language models enable few-shot clustering}.
\newblock \emph{Transactions of the Association for Computational Linguistics}, 12:321--333.

\bibitem[{Wang et~al.(2024)Wang, Yang, Huang, Yang, Majumder, and Wei}]{wang-etal-2024-improving-text}
Liang Wang, Nan Yang, Xiaolong Huang, Linjun Yang, Rangan Majumder, and Furu Wei. 2024.
\newblock \href {https://doi.org/10.18653/v1/2024.acl-long.642} {Improving text embeddings with large language models}.
\newblock In \emph{Proceedings of the 62nd Annual Meeting of the Association for Computational Linguistics (Volume 1: Long Papers)}, pages 11897--11916, Bangkok, Thailand. Association for Computational Linguistics.

\bibitem[{Zhang et~al.(2021)Zhang, Nan, Wei, Li, Zhu, McKeown, Nallapati, Arnold, and Xiang}]{zhang-etal-2021-supporting}
Dejiao Zhang, Feng Nan, Xiaokai Wei, Shang-Wen Li, Henghui Zhu, Kathleen McKeown, Ramesh Nallapati, Andrew~O. Arnold, and Bing Xiang. 2021.
\newblock \href {https://doi.org/10.18653/v1/2021.naacl-main.427} {Supporting clustering with contrastive learning}.
\newblock In \emph{Proceedings of the 2021 Conference of the North American Chapter of the Association for Computational Linguistics: Human Language Technologies}, pages 5419--5430, Online. Association for Computational Linguistics.

\bibitem[{Zhang et~al.(2019)Zhang, Basu, and Davidson}]{DBLP:conf/pkdd/ZhangBD19}
Hongjing Zhang, Sugato Basu, and Ian Davidson. 2019.
\newblock \href {https://doi.org/10.1007/978-3-030-46150-8\_4} {A framework for deep constrained clustering - algorithms and advances}.
\newblock In \emph{Machine Learning and Knowledge Discovery in Databases - European Conference, {ECML} {PKDD} 2019, W{\"{u}}rzburg, Germany, September 16-20, 2019, Proceedings, Part {I}}, volume 11906 of \emph{Lecture Notes in Computer Science}, pages 57--72. Springer.

\bibitem[{Zhang et~al.(2020)Zhang, Kishore, Wu, Weinberger, and Artzi}]{DBLP:conf/iclr/ZhangKWWA20}
Tianyi Zhang, Varsha Kishore, Felix Wu, Kilian~Q. Weinberger, and Yoav Artzi. 2020.
\newblock \href {https://openreview.net/forum?id=SkeHuCVFDr} {Bertscore: Evaluating text generation with {BERT}}.
\newblock In \emph{8th International Conference on Learning Representations, {ICLR} 2020, Addis Ababa, Ethiopia, April 26-30, 2020}. OpenReview.net.

\bibitem[{Zhang et~al.(2023)Zhang, Wang, and Shang}]{zhang-etal-2023-clusterllm}
Yuwei Zhang, Zihan Wang, and Jingbo Shang. 2023.
\newblock \href {https://doi.org/10.18653/v1/2023.emnlp-main.858} {{C}luster{LLM}: Large language models as a guide for text clustering}.
\newblock In \emph{Proceedings of the 2023 Conference on Empirical Methods in Natural Language Processing}, pages 13903--13920, Singapore. Association for Computational Linguistics.

\bibitem[{Zheng et~al.(2023)Zheng, Chiang, Sheng, Zhuang, Wu, Zhuang, Lin, Li, Li, Xing, Zhang, Gonzalez, and Stoica}]{10.5555/3666122.3668142}
Lianmin Zheng, Wei-Lin Chiang, Ying Sheng, Siyuan Zhuang, Zhanghao Wu, Yonghao Zhuang, Zi~Lin, Zhuohan Li, Dacheng Li, Eric~P. Xing, Hao Zhang, Joseph~E. Gonzalez, and Ion Stoica. 2023.
\newblock Judging llm-as-a-judge with mt-bench and chatbot arena.
\newblock In \emph{Proceedings of the 37th International Conference on Neural Information Processing Systems}, NIPS '23, Red Hook, NY, USA. Curran Associates Inc.

\end{thebibliography}
